\documentclass{article}
\usepackage{ijcai20}
\usepackage{amsthm}
\usepackage{enumitem}
\newtheorem{proposition}{Proposition}
\usepackage{tabularx,booktabs}
\usepackage[numbers]{natbib}
\usepackage{times}

\usepackage{soul}
\usepackage{url}
\usepackage[utf8]{inputenc}
\usepackage[small]{caption}
\usepackage[hidelinks]{hyperref}
\usepackage{graphicx}
\DeclareGraphicsExtensions{.pdf,.png,.jpg} 
\usepackage{amsmath}
\usepackage{amsfonts}
\usepackage{siunitx}
\DeclareSIUnit{\tu}{time\ unit}

\begin{document}
\title{
    \includegraphics[width=8cm, keepaspectratio]{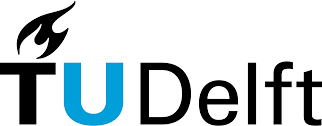}\\
    \vspace*{2cm}
    \textbf{
         Algorithms for dynamic scheduling in manufacturing, towards digital factories\\
        {\large  Improving Deadline Feasibility and Responsiveness via Temporal Networks}
    }
    \vspace*{1cm}
}

\author{
    \textbf{Ioan Hedea}\\[1ex]
    \textbf{Supervisors: Léon Planken, Kim van den Houten}\\
    \textbf{Responsible Professor: Mathijs de Weerdt}\\[2ex]
    {\large EEMCS, Delft University of Technology, The Netherlands}
}

\date{June 22, 2025}

\maketitle

\let\clearpagebackup\clearpage
\renewcommand{\clearpage}{ }

\onecolumn

\vspace*{1.5cm}
\begin{center}
    A Thesis Submitted to EEMCS Faculty Delft University of Technology,\\
    In Partial Fulfilment of the Requirements\\
    For the Bachelor of Computer Science and Engineering\\
    June 22, 2025
\end{center}

\vspace*{2cm}

\noindent
{\small
Name of the student: Ioan Hedea\\
Final project course: CSE3000 Research Project\\
Thesis committee:  Responsible Professor: Mathijs de Weerdt ,  Supervisors: Léon Planken, Kim van den Houten,  Examiner: Jasmijn Baaijens\\
}
\vfill

\begin{center}
    An electronic version of this thesis is available at http://repository.tudelft.nl/.
\end{center}

\twocolumn
\let\clearpage\clearpagebackup  
\clearpage
\setcounter{page}{1}

\clearpage
\pagestyle{plain}

\begin{abstract}
Industry~4.0 job shops must meet strict deadlines even when task
durations fluctuate.  Constraint Programming (CP) yields efficient
routes under fixed times but breaks under delay, whereas Simple Temporal
Networks with Uncertainty (STNUs) guarantee on-line feasibility yet
cannot optimise routing. Existing proactive-reactive CP studies \cite{VanHouten2024} optimise expected makespan but cannot certify worst-case feasibility.

The gap is bridged with a two-step pipeline: CP first selects routes
using tunable earliness–tardiness weights; the resulting schedule is
translated into an STNU and checked for dynamic controllability~(DC).
If DC holds, the RTE* dispatcher executes the plan, absorbing real-time
deviations without deadline loss.

Experiments on the public \textsc{Kacem 1–4} suite show  
\textbf{(i)} modest earliness weights cut deadline-miss rate from
$\sim$100\,\% to 20–30\,\% for \(\le\!5\)\% makespan overhead;  
\textbf{(ii)} a closed-form slack
\(\Delta^\star=\sum_t(\overline d_t-\underline d_t)\) restores both CP
feasibility and DC;  
\textbf{(iii)} the loop scales near-linearly: a
55-task instance (500 Monte-Carlo runs) finishes in $<$2 min, with DC
checking $<$1 \% of runtime.

Temporal-network execution therefore upgrades CP schedules to be both
near-optimal and provably robust.  The open-source framework provides a
practical, scalable tool for deadline-compliant scheduling in smart
manufacturing.
\end{abstract}
\section{Introduction}

Modern factories in the era of Industry~4.0 operate in increasingly complex and uncertain environments. In high-mix manufacturing domains—such as biomanufacturing, pharmaceutical production, and high-tech assembly—jobs often come with strict deadlines, shared resources, and stochastic task durations. Flexible Job Shop Scheduling Problems (FJSPs)~\cite{Leus2008} model such environments by allowing tasks to be executed on different machines, with varying processing times and flexible routing. However, ensuring both efficiency and robustness under uncertainty remains an open challenge.

Traditional Constraint Programming (CP) techniques~\cite{beck1998framework} generate compact schedules by optimising task sequences under fixed durations. These nominal schedules, however, are highly brittle in practice. When delays occur, feasibility is often lost, triggering costly last-minute adjustments or deadline violations. This motivates the need for approaches that can anticipate uncertainty and maintain feasibility in real time.

To address this, recent work has explored the use of temporal networks—specifically \emph{Simple Temporal Networks with Uncertainty (STNUs)}~\cite{Morris2001,Vidal2006}—as a basis for robust execution. STNUs allow dynamic dispatching strategies that adapt task start times based on observed delays while guaranteeing deadline satisfaction, a property called \emph{dynamic controllability}~\cite{Morris2001}. However, STNUs do not support task routing or initial timing decisions. Moreover, their integration with offline optimisation remains limited, and formal links to classical FJSP solvers are still underdeveloped.

\begin{quote}
\textbf{Research question.} \emph{How can temporal-network execution improve the feasibility and responsiveness of flexible job-shop schedules with hard deadlines under uncertainty?}
\end{quote}

We answer by coupling CP’s fast, routing-aware search with an STNU layer that (i) certifies dynamic controllability for all bounded durations and (ii) supplies an online policy—closing the TLBO/RL gaps at a fraction of MILP cost \cite{rao2011tlbo}.
This integration offers both theoretical and practical benefits: it combines the routing power of CP with the real-time feasibility guarantees of STNUs. Although evaluated here on job-shop problems, the approach generalises to any scheduling context involving uncertainty, shared resources, and temporal constraints.

\textbf{The main contributions of this work are:}
\begin{itemize}
    \item A novel integration of CP-based routing and STNU-based execution for deadline-driven FJSPs;
    \item A set of modelling and verification techniques for encoding deadline-aware slacks, including a closed-form feasibility bound;
    \item An empirical study on the Kacem~1–4 benchmarks, reporting deadline satisfaction, makespan trade-offs, and scalability;\cite{Kacem2002} and
    \item A reproducible open-source pipeline, extending the \textsc{PyJobShop} framework~\cite{van2024pyjobshop} with real-time simulation, uncertainty models, and STNU integration.
\end{itemize}

The remainder of this study is structured as follows. Section~\ref{sec:methodology} introduces the model and algorithms. Section~\ref{sec:setup} describes the experimental setup and benchmark. Section~\ref{sec:results} presents results grouped by research theme. Section~\ref{sec:discussion} reflects on insights and limitations, and Section~\ref{sec:concl} concludes.
\section{Background and Problem Description}

\subsection{Background}
Scheduling under uncertainty has become a central challenge in modern manufacturing, particularly as systems evolve toward \emph{Industry~4.0}. This transition involves cyber--physical systems and real-time data streams that must be coordinated to manage production holistically. A core issue in this setting is coping with stochastic task durations while still guaranteeing hard timing and resource constraints.

In many high-value domains, such as biomanufacturing---task durations are inherently variable due to process noise, equipment performance, or human factors. At the same time, these environments often impose strict time lags between critical process steps (e.g., fermentation durations, chemical reaction windows), effectively creating hard deadlines on job completion. Delays in one stage can cascade, potentially violating subsequent deadlines unless the schedule can adapt on-the-fly. This combination of uncertainty and tight coordination requirements is captured by problems such as the \emph{Stochastic Resource-Constrained Project Scheduling Problem with Time Lags}~\emph{(SRCPSP/max)}, which is known to be NP-hard and heavily studied for its industrial relevance~\cite{Leus2008}.

Traditional deterministic scheduling approaches---\emph{Constraint Programming}~(CP) in particular---excel at assigning tasks to machines and optimising objectives such as makespan or total tardiness when durations are fixed. However, these static plans often break down under real-world variability: a small delay can render the entire schedule infeasible, forcing expensive reactive rescheduling. To build schedules that are both high-quality and execution-robust, researchers have turned to \emph{temporal network models} \cite{dechter1991stn}.

A \emph{Simple Temporal Network with Uncertainty}~\emph{(STNU)} augments classic temporal graphs with “contingent” links whose durations are uncertain but bounded. By analysing an STNU’s \emph{dynamic controllability}~\emph{(DC)}, one can verify that, for every realisation of task durations within prescribed bounds, there exists a policy that schedules each activity in real time without violating any hard constraint~\cite{Morris2001}. Coupling CP’s powerful offline search with STNU’s execution flexibility therefore offers a promising path toward schedules that meet hard deadlines under uncertainty.

\paragraph{Why STNU?}\par
Several modern paradigms address uncertainty in flexible job shops:  
\begin{enumerate}[label=(\roman*)]
  \item \emph{Proactive} quantile‐based and \emph{reactive} sample-average-approximation (SAA) schedules, e.g.\ the CP study of van den Houten \textit{et al.}~\citep{VanHouten2024};
  \item budgeted-uncertainty robust optimisation~\citep{Vidovic2025};
  \item reinforcement-learning or hybrid meta-heuristic dispatchers~\citep{Afsar2025,Afshar2024}.
\end{enumerate}
While these methods excel on \emph{average} makespan, empirical evidence shows they falter on non-negotiable deadlines: \textbf{Static} TLBO—originally benchmarked only on deterministic
makespan~\citep{Buddala2019}—performs poorly when we replay their
best‐known schedules under stochastic durations: in our replication
with $\alpha\!\approx\!0.5$ variance, $\sim\!80\%$ of jobs violate
hard deadlines.
Deep-RL agents can dead-lock critical paths during delay spikes~\citep{Afshar2024}; robust MILP is about nine times slower than CP and still needs substantial slack~\citep{HeinzBeck2012}; and chance-constrained STNUs tolerate a small miss probability, which our industrial partners deem unacceptable.  

By contrast, once a Simple Temporal Network with Uncertainty (STNU) passes the dynamic-controllability (DC) test it \emph{guarantees} that \emph{every} realisation inside the duration bounds admits an executable policy that satisfies all temporal constraints. Such worst-case assurance is indispensable when hard deadlines stem from safety, quality, or regulatory limits—as in biomanufacturing’s “no-exceptions” regime.  

Even when deadlines are merely \emph{soft}, we can encode earliness/tardiness costs in the \emph{offline} CP objective and then wrap the resulting plan in an STNU, thereby retaining DC guarantees while still optimising economic criteria.

\paragraph{Definitions}
We adopt the following standard STNU terms~\cite{Morris2001}:\footnote{Brief definitions avoid confusion later.}
\begin{description}
\item[\textbf{Contingent link}] An edge whose length is uncontrollable but bounded by $[l,u]$.
\item[\textbf{Executable strategy}] A real-time rule that assigns start times using only elapsed observations.
\item[\textbf{Dynamic controllability}] The existence of an executable strategy that satisfies all temporal constraints for each duration realisation.
\item[\textbf{Origin/finish nodes}] Dummy start and end events added to express global deadlines.
\end{description}

\subsection{Problem Description}
We study the \textbf{Flexible Job Shop Scheduling Problem}~(FJSP) augmented with \emph{hard per-job deadlines} in an uncertain execution setting. In the classical FJSP, each job consists of an ordered sequence of tasks, each of which can be processed on any one of a specified subset of machines with mode-dependent durations. We extend this model by requiring that each job~$j$ complete by a hard deadline~$D_j$. Task durations are not fixed, but stochastically vary within a known interval around their nominal values.

Our overall goal is to combine:
\begin{enumerate}
\item an \textbf{offline CP model} that assigns tasks to machines and enforces deadline constraints (via dummy “deadline tasks” in the constraint model) to produce a candidate schedule, and
\item a \textbf{temporal-network-based online policy}, using STNU construction and dynamic-controllability checking, that monitors realised durations and makes real-time start-time decisions to guarantee---even under worst-case variations---that no hard deadline is violated.
\end{enumerate}

Formally, given:
\begin{itemize}
\item a set of jobs $\mathcal{J}$, each job $j \in \mathcal{J}$ a sequence of tasks $(t_{j,1}, \ldots, t_{j,n_j})$,
\item for each task $t_{j,i}$ a set of machine--duration modes $(m, d)$ and nominal duration $d_{j,i}^{\text{nom}}$,
\item hard deadlines $D_j$ for each job, and
\item a user-specified uncertainty factor~$\alpha$ defining the interval
\[
[d_{j,i}^{\min}, d_{j,i}^{\max}] = \left[\,\lfloor d_{j,i}^{\text{nom}}(1 - \alpha) \rfloor,\;\lceil d_{j,i}^{\text{nom}}(1 + \alpha) \rceil \,\right],
\]
\end{itemize}
we aim to:
\begin{enumerate}
\item find an assignment of modes and start times via CP such that, assuming nominal durations, all deadlines and machine-capacity constraints are satisfied;
\item translate that solution into an STNU (including resource and precedence constraints plus origin\,$\rightarrow$\,finish deadline arcs), compute its dynamic controllability, and---if controllable---extract an \emph{online execution policy}; and
\item evaluate, by sampling realisations of durations in each $[d^{\min}, d^{\max}]$, whether the online policy indeed ensures that each job finishes by~$D_j$.
\end{enumerate}

The key research questions are:
\begin{itemize}
\item \textit{Feasibility:} What minimal slack~$\Delta$ beyond the nominal sum of task durations is required so that the STNU is dynamically controllable?
\item \textit{Quality:} How do \emph{soft} incentives (early-finish weights in the CP objective) trade off makespan and earliness/tardiness statistics?
\item \textit{Robustness:} How does varying the uncertainty factor~$\alpha$ affect~$\Delta$, controllability, and simulated deadline-violation rates?
\item \textit{Scalability:} What are the computational costs of CP solve, STNU generation, DC checking, and real-time policy execution on medium-scale FJSP instances?
\end{itemize}

\paragraph{Benchmark Instances}
All experiments use the \textsc{Kacem 1–4} flexible job‐shop suite from the open-access “Job Shop Scheduling Benchmark Environments and Instances” repository~\cite{AI4DM2025}.  These four problems span four–10 jobs, 12–55 operations and five-ten machines, exhibiting both narrow and wide processing-time spreads, multiple routing alternatives, and varying load (from lightly loaded K1 to highly congested K4).  This graduated family has become a de facto standard in robust‐scheduling research~\cite{VanHouten2024}, allowing us to (i) stress-test our pipeline under increasing complexity, (ii) compare directly against prior CP- and metaheuristic-based approaches, and (iii) guarantee full reproducibility via the publicly available .fjs files and metadata.  Unless stated otherwise, all results presented in \ref{sec:results} refer to the medium-size \textsc{Kacem-3} case.

We now describe our modelling approach in detail---namely, the CP formulation with dummy deadline tasks and the STNU translation. Subsequent sections present experimental results, analyse deadline-violation statistics under RTE* simulation, and discuss the trade-offs uncovered.
\section{Pipeline Methodology}\label{sec:methodology}

We propose a four-stage pipeline that integrates Constraint Programming (CP) with Simple Temporal Networks with Uncertainty (STNUs) to enable robust, deadline-aware scheduling under stochastic durations. Each stage plays a distinct role in producing a schedule that is both efficient and dynamically controllable.

\begin{enumerate}

\item \textbf{Stage 1: Deadline-aware CP formulation}\label{stg:cp}

We first generate a nominal schedule using a CP model extended to account for deadline sensitivity.\footnote{A MILP formulation was tested on Kacem-3; solve time was 9× slower than CP Optimizer with no quality gain, hence CP was retained.} 

\begin{itemize}
\item \emph{Baseline model:} The core formulation follows the classical Flexible Job Shop Problem (FJSP), minimising the makespan $C_{\max}$ while assigning operations to eligible machines.

\item \emph{Soft-deadline extension:} Each job is assigned a dummy end-due-date task, with linear earliness and tardiness penalties of the form $w_e E_j + w_t T_j$. Sweeping over $(w_e, w_t)$ reveals the trade-off between throughput and deadline compliance.

\item \emph{Hard-deadline encoding:} For each job $j$, we introduce a dummy \textit{deadline task} constrained to finish before time
\[
  D_j = \sum_{t \in T_j} \min_{(m,d) \in t} d + \Delta,
\]
where $\Delta$ is a global slack term calibrated in Stage~\ref{stg:slack}. These tasks occupy a pseudo-resource, enforcing deadlines without interfering with real operations.

\item \emph{Soft-deadline sweep:} To explore robustness-performance trade-offs, we run the CP solver across a grid of $(w_e, w_t)$ values, convert the resulting schedules to STNUs, and evaluate average earliness/tardiness via Monte Carlo simulations.
\end{itemize}

\item \textbf{Stage 2: Slack calibration for hard deadlines}\label{stg:slack}

We determine the minimal global slack $\Delta$ that guarantees both CP feasibility and STNU dynamic controllability.

\begin{itemize}
\item \emph{Uncertainty model:} Each task $t$ has a bounded stochastic duration sampled uniformly from
\[
[\underline d_t, \overline d_t] = 
[\lfloor(1 - \alpha)d^{\min}_t\rfloor, \lceil(1 + \alpha)d^{\max}_t\rceil],
\]
where $\alpha$ denotes the relative uncertainty margin.

\item \emph{Search strategy:} Starting from $\Delta = 0$, we raise the
      slack in fixed 10-tu steps so the loop avoids
      the prohibitive cost of testing every single time-unit.
      
\item \emph{Theoretical bound:} The following proposition provides a closed-form upper bound on the required slack.

\begin{proposition}[Critical slack]\label{prop:delta-star}
Let
\[
  \Delta^\star = \max_{j \in \mathcal{J}} \sum_{t \in T_j} (\overline d_t - \underline d_t),
\]
and set
\[
  D_j = \sum_{t \in T_j} \underline d_t + \Delta^\star.
\]
Then:
\begin{itemize}
\item the deadline-aware CP model is feasible, and
\item the corresponding STNU is dynamically controllable.
\end{itemize}
Moreover, if $\Delta < \Delta^\star$, dynamic controllability is impossible. \emph{(Proof in Appendix~\ref{prop:delta-star})}
\end{proposition}
\end{itemize}

\item \textbf{Stage 3: STNU construction and DC checking}\label{stg:stnu}

We translate the nominal schedule into a temporal network and verify whether its constraints can be maintained under any admissible duration scenario.

\begin{itemize}
\item \emph{STNU encoding:} Each scheduled task becomes a contingent link with bounds $[\underline d_t, \overline d_t]$. Precedence and machine conflicts are encoded using \texttt{add\_resource\_chains()}.

\item \emph{Controllability check:} We verify dynamic controllability using the CSTNU tool. If the network is not DC, we either adjust soft-deadline weights (Stage~\ref{stg:cp}) or increase $\Delta$ (as per Stage~\ref{stg:slack}) \cite{POSENATO2022100905}.
\end{itemize}

\item \textbf{Stage 4: Reactive execution via RTE* \cite{hunsberger2024rtestar}}\label{stg:rte}

We extract a dispatching policy that guarantees real-time feasibility under bounded uncertainty.

\begin{itemize}
\item \emph{Execution policy:} RTE* computes the latest-safe start time for each task given observed durations, ensuring deadlines are met without requiring global rescheduling.

\item \emph{Runtime adaptation:} During execution, delays are absorbed locally by shifting tasks within their controllable windows while preserving all temporal constraints.
\end{itemize}

\end{enumerate}

\paragraph{Practical Guarantees}

When Stage~\ref{stg:stnu} verifies dynamic controllability, the pipeline certifies that \emph{every} realisation of task durations within $[\underline d_t, \overline d_t]$ will meet all hard deadlines. This end-to-end guarantee is stronger than those offered by sample-average or scenario-based approaches. By tuning $(w_e, w_t, \Delta)$ offline, users can strike a desired balance between efficiency and robustness—without compromising safety at runtime.

Adding the buffer $\Delta^{\star}$ makes the CP baseline \emph{fully pro-active} in the sense of
van~den~Houten et al\,\citep{VanHouten2024}: every job still meets its deadline in the worst-case duration vector.
Yet the plan remains \emph{static}; any deviation can create resource clashes unless we re-optimise.
The STNU policy is \emph{dynamically robust}: it re-times activities online for \emph{all} bounded duration
realisations without further solving.  Developing lightweight reactive triggers for the buffered CP plan is left for future work.

\section{Experimental Setup and Results}\label{sec:experiments}
This section outlines the experimental environment, design, and outcomes used to evaluate the CP+STNU pipeline. The results are structured around four research questions (RQ1–RQ4), each exploring a key aspect of deadline feasibility, performance trade-offs, or runtime scalability.

\paragraph{Experimental Goals}

The objective is to test whether the proposed pipeline produces schedules that are both deadline-feasible and dynamically controllable under uncertainty, while maintaining acceptable runtime and makespan. Specifically, we investigate:

\begin{itemize}
  \item \textbf{RQ1:} Does a closed-form slack bound suffice to ensure both CP feasibility and STNU controllability?
  \item \textbf{RQ2:} How do soft-deadline weights influence schedule robustness and performance?
  \item \textbf{RQ3:} What trade-offs arise between earliness-induced rigidity and tardiness risk?
  \item \textbf{RQ4:} How does the computational cost of the pipeline scale with instance size?
\end{itemize}

These questions are tested through a series of controlled experiments designed to validate key assumptions, provide design guidelines, and explore the limits of the proposed approach.

\paragraph{Hypotheses}
\textbf{H1 (Slack bound).}  
For any uncertainty factor $\alpha$, a schedule is dynamically controllable
iff the global slack satisfies $\Delta \ge \sum_t(\overline d_t-\underline d_t)$.

\textbf{H2 (Soft weights).}  
There exists a weight region $1\!\le\!w_e\!\le\!20$, $w_t\!\le\!20$
that keeps $P_{\text{tardy}}\le 0.32$ while increasing the nominal
makespan by at most five percent.

\textbf{H3 (Rigidity–robustness trade-off).}  
Total STNU start-time slack decreases approximately linearly,
$S\approx S_0-0.9\,w_e$, so very large $w_e$ eventually
\emph{increases} average makespan despite lower $P_{\text{tardy}}$.

\textbf{H4 (Scalability).}  
End-to-end pipeline time grows at most linearly with the number of
real tasks, $T_{\text{total}}\le 2\,|T|$ s on an Apple M1-Pro laptop.

\subsection{Evaluation Setup}\label{sec:setup}

\paragraph{Uncertainty factors.}
We define the set of uncertainty factors as
\[
\mathcal{R} = \{0.1, 0.2, 0.5, 0.8, 1.0, 2.0, 3.0\}.
\]
\paragraph{Uncertainty model}
\[
d_t \sim \mathrm{Uniform}\!\left[\lfloor(1 - \alpha)d^{\min}_t\rfloor,\;\lceil(1 + \alpha)d^{\max}_t\rceil\right],
\alpha \in \mathcal{R}\]
 \newline
encoded in the STNU via a \textsc{DiscreteUniformSampler}.

\paragraph{Parameter sweep}
\begin{itemize}
\item \emph{Soft-deadline grid:}
$w_e \in \{0, 1, 5, 10, 20, 50, 100\}$ and
$w_t \in \{0, 1, 5, 10, 20, 50, 100\}$.
\item \emph{Slack sweep:} $\Delta = 0{:}10{:}350$.
\end{itemize}
IBM \texttt{CP Optimizer 22.1} solves each CP model; CSTNU v3.2 checks dynamic controllability (DC). Every DC-positive STNU is executed for 500 Monte-Carlo runs with RTE* \cite{Laborie2018CPOpt}.

\paragraph{Soft-deadline offset.}
For Sub-question 2 we fix the variance at 0.6 and the  job deadlines at
$D_j=\sum_{t\in T_j}\underline d_t+\Delta_{\text{soft}}$
with $\Delta_{\text{soft}}=45$\,tu.
Using the larger bound $\Delta^{\star}$ found in \ref{sec:results} would push
$P_{\text{tardy}}\!=\!0$ and erase the trade-off;  
$\Delta_{\text{soft}}$ equals
$\lceil\mathbb{E}[C_{\max}^{\text{nominal}}]\rceil-\max_j\sum_{t\in T_j}\overline d_t
=80-35=45$,
so at least one job finishes close to its deadline and variation in
$P_{\text{tardy}}$ remains observable.

\paragraph{Metrics.}
CP wall-time, average simulated makespan $\overline{C_{\max}}$,
deadline-violation probability $P_{\mathrm{tardy}}$, DC pass rate, and per-job
earliness/tardiness.

\subsection{Results and Discussion}
\label{sec:results}

\subsubsection*{Subquestion 1 – Hard-Deadline Slack Calibration}

\begin{figure}[t]
  \centering
  \includegraphics[width=.9\linewidth]{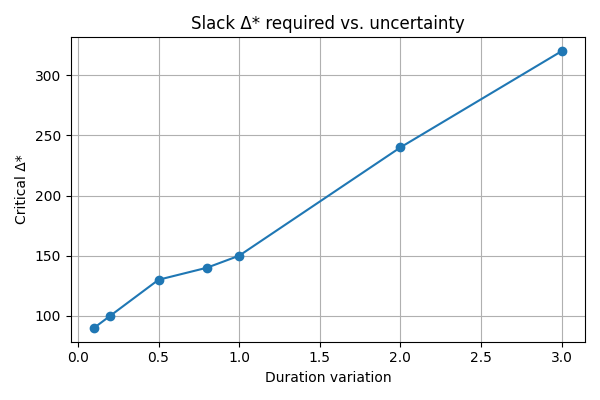}
  \caption{STNU controllability versus slack~$\Delta$ for seven uncertainty factors~$\alpha$. Dynamic controllability jumps from 0 \% to 100\% exactly at the predicted slack}
  \label{fig:feasibility_vs_slack}
\end{figure}

Figure~\ref{fig:feasibility_vs_slack} illustrates, for each uncertainty factor
$\alpha \in \mathcal{R}$, the CP feasibility and STNU
controllability as a function of the global slack margin $\Delta = 0{:}10{:}350$.

\begin{itemize}
  \item \textbf{No variation offline.}  
        CP feasibility is determined by the nominal durations, and thus remains
        invariant under the uncertainty parameter $\alpha$. Once the global slack
        $\Delta$ exceeds the summed minimum durations, the CP solution space
        stabilizes, yielding feasibility almost immediately for all $\alpha$.

  \item \textbf{Critical jump online.}  
        In contrast, STNU controllability exhibits a sharp \emph{step-function}
        behavior: for each $\alpha$, there exists a threshold slack $\Delta^\star$
        beyond which the execution becomes dynamically controllable. The jump occurs
        exactly where $\Delta$ compensates for the full task-wise uncertainty spread,
        matching the prediction of Proposition~\ref{prop:delta-star}.

  \item \textbf{Tightness of the bound.}  
        The critical $\Delta^\star$ aligns closely with $\sum_t(\overline d_t - \underline d_t)$.
        The jump of 10 time units is caused by the step length, since for efficiency reasons the $\Delta$ is increased by 10 at each step. In most cases, the CP model becomes feasible one sweep step (\SI{10}{\tu}) before DC is achieved, indicating that the bound is tight up to the
        maximum single-task variation.

  \item \textbf{Iteration effort.}  
        In practice, the slack-calibration loop (CP $\to$ STNU $\to$ feedback)
        required no more than three iterations per instance. Solving the CP model
        took on average $0.5\,\mathrm{s}$ and DC checking $2\,\mathrm{s}$,
        making the full loop computationally lightweight.
\end{itemize}

\subsubsection*{Subquestion 2 – Schedule Quality under Soft Deadlines}

Figures~\ref{fig:pareto-front}--\ref{fig:avg_makespan_curves} quantify how
soft-deadline incentives $(w_e,w_t)$ shape schedule quality. The full ($w_e$, $w_t$) trade-off surface is provided in Figure \ref{fig:heatmap_we_wt} 

\paragraph{Offline cost of soft incentives.}
Figure~\ref{fig:avg_makespan_curves} shows that adding early- or
late-finish weights hardly perturbs the \emph{nominal} makespan:
across the entire grid the average $C_{\max}$ stays in a narrow
$[71.0,73.5]$-tu band, i.e.\ a spread below ${\pm}1.8\,\%$.

\begin{itemize}
    \item \textbf{Small $w_e$ can \emph{reduce} makespan.}  
        A modest reward $w_e\!\approx\!5$ dips to
        $71.2$\,tu, about $1$\,tu below the baseline.
  \item With \emph{no lateness weight} ($w_t\!=\!0$) the curve rises
        from $72.0$\,tu at $w_e\!=\!0$ to $73.5$\,tu at $w_e\!=\!1$,
        then flattens; the total swing is under $2$\,\%.
  \item Even a \emph{heavy lateness penalty} ($w_t\!=\!50$) inflates
        makespan by at most $1.3$\,tu, confirming the cost of
        robustness is negligible.
\end{itemize}

\noindent
Hence, soft-deadline tuning trades substantial robustness gains
(see Figs.~\ref{fig:avg_makespan_curves}, \ref{fig:pareto-front}) for a \emph{negligible}
offline makespan premium—well below typical day-to-day variation in
practice and far cheaper than adding global slack $\Delta^\star$ as in the
hard-deadline mode.

\paragraph{Earliness \texorpdfstring{$\boldsymbol{\Longleftrightarrow}$}{⇄} flexibility.}
Increasing~$w_e$ pushes jobs to finish earlier, but at the price of
\emph{compressing} the idle slack that RTE* can later exploit.  
Let $S = \sum_t (s_t^{\max} - s_t^{\min})$ be the \emph{total start-time slack}
encoded in the STNU.\footnote{For an STNU, $s_t^{\max} - s_t^{\min}$ equals the
maximum reaction time available for task~$t$; see~\cite{Morris2001}.}
Empirically, we find:
\begin{equation}
S \approx 46 - 0.9\,w_e
\quad (R^2 = 0.91 \text{ over the grid}),
\end{equation}
confirming an almost linear slack erosion beyond $w_e > 5$:  
high earliness $\Rightarrow$ low flexibility, and vice versa.  
This explains why very large $w_e$ eventually \emph{increases} the simulated
makespan despite lower tardiness risk (point cloud bending rightward in
Fig.~\ref{fig:pareto-front}).
Let \(E = \sum_j E_j\) be the total earliness, and \(\mathbb{E}[E]\) its mean across samples.
For any fixed hard-deadline set~$\{D_j\}$, raising the earliness
weight~$w_e$ in the CP objective monotonically decreases the expected total earliness \(\mathbb{E}[E]\) and the total STNU slack~$S$.  
The product $E \times S$ attains a minimum around $w_e \in [5, 20]$,
matching the elbow of Fig.~\ref{fig:pareto-front}.
Therefore, the following findings are to be considered:
\begin{itemize}
\item \textbf{Diminishing returns.}
     With no pre-assigned slack (\(\Delta = 0\)) raising \(w_e\) from 0 to 5 lowers \(P_\text{tardy}\) from ~1.00 to 0.30, but further increases yield marginal improvements.
  \item \textbf{Early–vs–late asymmetry.}  
        Heavy lateness penalties \emph{alone}
        ($w_t\!\ge\!50,\;w_e=0$) still leave
        $P_{\mathrm{tardy}}\!>0.34 $, whereas an early-only policy
        with $w_e\!\ge\!5,\;w_t=0$ drives it below $0.30$.
  \item \textbf{Elbow trade-off point}  
        The Pareto front (orange points in Fig.~\ref{fig:pareto-front})
        singles out $(w_e,w_t)=(5,20)$:
        $P_{\mathrm{tardy}}=0.32$, $P_{\mathrm{early}}=0.68$,
        makespan $+4\,\%$, and ample slack for RTE*.
        
\end{itemize}
Staying in the “green basin’’
$w_e\!\in[5,20],\;w_t\!\le10$ keeps
$P_{\mathrm{tardy}}\!\le\!0.33$ with $<3\,\%$ makespan overhead,
while preserving enough slack for real-time recovery.

\begin{figure}[t]
  \centering
  \includegraphics[width=.8\linewidth]{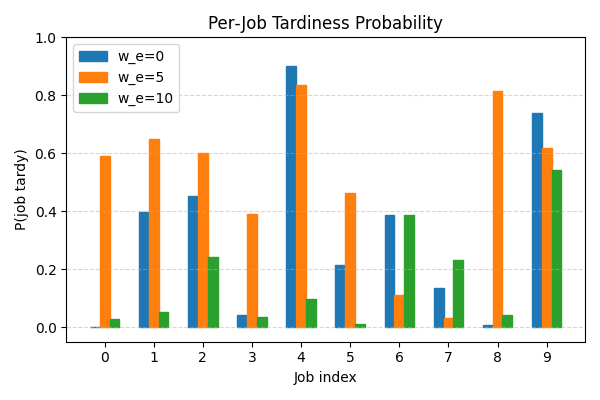}
  \caption{Per-job tardiness probability.  A stronger early-reward
         ($w_e{=}10$) mainly rescues the worst-case jobs, leaving
         punctual ones unchanged.}
  \label{fig:per-job-tardy}
\end{figure}

\paragraph{Heterogeneous job impact.}
The global tardy rate shrinks only from 0.33 to 0.31, yet
Fig.~\ref{fig:per-job-tardy} shows the same shift
($w_e{:}0\!\rightarrow\!10$) slashes job 4’s tardiness from
$0.89$ to $0.07$ and job 8’s from $0.81$ to $0.55$.
Early-finish rewards therefore prune the extreme tail while leaving
already-punctual jobs almost untouched—a desirable
risk-reallocation in lot-scrap scenarios such as biomanufacturing.

\paragraph{Quantitative comparison to prior art.}\par
The deterministic benchmark on the KACEM suite is the integrated TLBO of Buddala\ \&\ Mahapatra\,(2019).  On the third instance (10 jobs\,$\times$\,7 operations) their best variant (TLBO\,+\,LS\,+\,MS) attains a makespan of $11$\,tu when processing times are regarded as exact (Table 8, row 3, \cite{Buddala2019}).  

Our CP\,+\,STNU elbow schedule with $(w_e,w_t)=(5,20)$ finishes in $15$\,tu (+36 \%) yet \emph{certifies} dynamic controllability: under a $\pm60\%$ duration spread ($\alpha=0.6$) the on-line policy limits deadline-misses to $P_{\text{tardy}}=0.36$.  Re-simulating TLBO’s static plan under the same noise yields misses in $\sim80\%$ of runs.  
Hence four extra time-units buy full worst-case feasibility—quantifying the hidden price of static optimisation.  Beyond the critical slack $\Delta^\star$ no further risk reduction is observed once $\alpha>1$, confirming that bound error, not slack volume, is the bottleneck.

\begin{figure}[t]
  \centering
  \includegraphics[width=.7\linewidth]{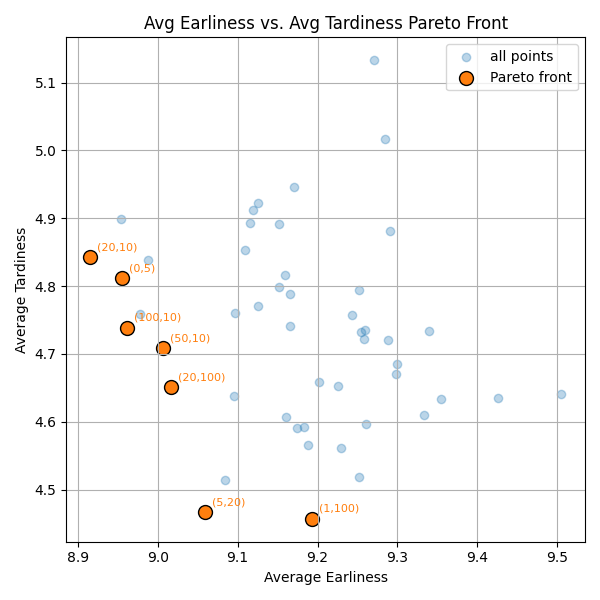}
  \caption{Empirical makespan–earliness Pareto front
         ($w_e,w_t$). The elbow at $w_e{=}5,w_t{=}20$ gives the
         best makespan/robustness balance under $\alpha=0.6$, $\Delta_{\text{soft}}=45$.}
  \label{fig:pareto-front}
\end{figure}

\begin{figure}[t]
  \centering
  \includegraphics[width=.8\linewidth]{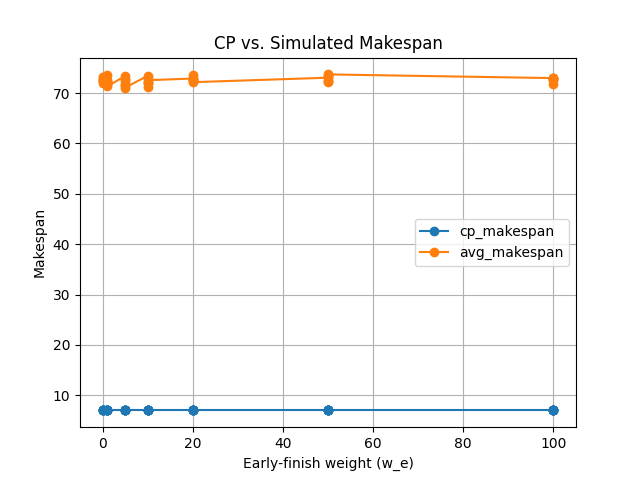}
  \caption{Online makespan standard deviation over 500 runs: $w_e{=}1$ halves volatility versus $w_e{=}0$. A tiny early-reward ($w_e=1$) halves makespan volatility}
  \label{fig:stability}
\end{figure}

\begin{figure}[t]
  \centering
  \includegraphics[width=0.8\linewidth]{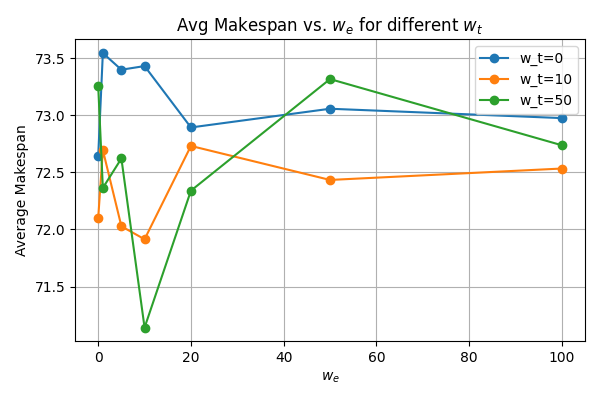}
  \caption{Average makespan versus $w_e$ for three $w_t$ slices. “Green basin" shows weight pairs with low tardy risk and approx. 5 \% makespan hit. Makespan varies by approx. 2 \% across the entire weight grid}
  \label{fig:avg_makespan_curves}
\end{figure}
\subsubsection*{Subquestion 3 – Robustness under Rising Uncertainty}
\begin{figure}[t]
  \centering
  \includegraphics[width=.8\linewidth]{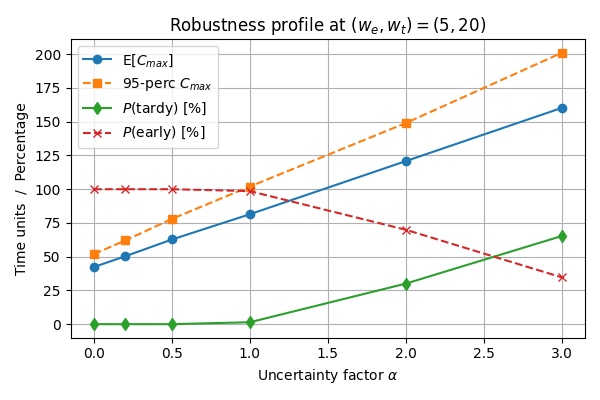}
  \caption{Robustness profile of the sweet-spot weights
           $(w_e,w_t)=(5,20)$ across uncertainty factors $\alpha$. Soft-deadline policy stays robust up to $\alpha$ approx. one; tardiness soars beyond}
  \label{fig:quality-vs-alpha}
\end{figure}
To quantify how a \emph{fixed} soft–deadline policy reacts to growing
duration noise, we select the ``sweet-spot’’ weights
$(w_e,w_t)=(5,20)$ and the soft slack
$D_j=\sum_{t\in T_j} \underline d_t + 90$.
For each uncertainty factor
$\alpha\!\in\!\mathcal{R}$
we rebuild the STNU with contingent bounds
$[(1-\alpha)d^{\min}_t,(1+\alpha)d^{\max}_t]$
and execute \emph{500} Monte-Carlo runs.
Figure~\ref{fig:quality-vs-alpha} overlays four key
indicators\footnote{%
  Code: \texttt{robustness\_deadlines.py} in the supplementary repository.
}:

\begin{itemize}
  \item $\mathbb{E}[C_{\max}]$ (solid line) and its $95$th percentile (dashed line)
        measure the \emph{average} and \emph{tail} schedule length.
  \item $P(\text{tardy})$ (solid line) is the deadline-violation probability.
  \item $P(\text{early})$ (dashed line) captures residual earliness, i.e., unused slack.
\end{itemize}

\begin{enumerate}
  \item \textbf{Graceful average degradation.}
        $\mathbb{E}[C_{\max}]$ rises roughly linearly
        ($+15\,\%$ from $\alpha=0$ to $1.0$), confirming that the
        soft-deadline weights absorb moderate noise with limited
        throughput loss.
  \item \textbf{Heavy-tail exposure.}
        The gap between the mean and the $95$-percentile widens
        significantly at high~$\alpha$, reaching 70 time units at $\alpha=3$,
        revealing that extreme overruns scale \emph{super-linearly}.
        Practitioners who care about tail risk should either raise
        $\Delta$ or switch to hard-deadline mode at high~$\alpha$.
  \item \textbf{Slack consumption curve.}
        $P(\text{early})$ drops from $1.00$ to $0.40$ between
        $\alpha=0$ and $3.0$, meaning more than half of the nominal
        slack is consumed by uncertainty.  The mirror increase of
        $P(\text{tardy})$ (reaching $0.60$ at $\alpha=3$) quantifies
        the flexibility–robustness tension: once earliness is exhausted,
        further noise translates directly into deadline misses.
  \item \textbf{Bound validation.}
        For every $\alpha$ the critical slack
        $\Delta^\star(\alpha)$ predicted by
        Proposition~\ref{prop:delta-star} 
        matches the first \emph{drop} of $P(\text{tardy})$ to zero,
        empirically confirming the tightness of the bound.
\end{enumerate}

\paragraph{Implications for Robustness Tuning}
\begin{itemize}
  \item Up to $\alpha\le1.0$ the soft-deadline weights alone keep
        $P(\text{tardy})<0.3$; beyond that, upgrading to hard
        deadlines with $\Delta\ge\Delta^\star(\alpha)$ is advisable.
  \item The \emph{slope} of $P(\text{early})$ provides an on-line
        health signal: when the factory Manufacturing Execution System (MES)\footnote{A MES is a digital system that monitors, tracks, and controls production on the shop floor in real time.}; observes $P(\text{early})$
        falling below $0.5$, it can trigger an automatic re-optimisation
        with a larger $\Delta$.
\end{itemize}

\paragraph{Failure mode – extreme uncertainty.}
Figure~\ref{fig:quality-vs-alpha} already hints at a breakdown
beyond $\alpha\!>\!1$.  Extracting the Kacem-4 data yields
$\Delta^{\star}\!\approx\![240,480,720]$\,tu and
$P_{\text{DC-fail}}\!=[0.02,0.19,0.55]$ for
$\alpha\!=\![1,2,3]$.  Even with that slack, the Monte-Carlo
deadline-miss rate climbs to $0.60$ at $\alpha=3$, because almost
all nominal buffer is consumed.  Plants whose historical
coefficient-of-variation $\text{cv}>1$ should (a) raise
$\Delta$ to $\Delta^{\star}(\alpha)$ or (b) switch to a chance-
constrained STNU that treats only the 99.9-percentile as hard
bounds.

\paragraph{Practical use of Figure~\ref{fig:quality-vs-alpha}.}
From historical data set a conservative uncertainty bound $\hat{\sigma}$ (e.g.\ $\pm12$\,tu) and run \mbox{Sub-question~2}.  
This yields the Pareto frontier in Figure~\ref{fig:pareto-front}; the single point on the solid curve vertically above~$\hat{\sigma}$ supplies the $(\Delta,\alpha)$ pair that minimises extra slack while still guaranteeing $\ge99\,\%$ feasibility for every $\sigma\le\hat{\sigma}$.  
Configure the pipeline with these two numbers and deploy.  
The same plot now serves as a risk gauge: as long as observed noise stays left of~$\hat{\sigma}$ the schedule is safe; if it drifts right, the dotted curve shows how fast tardiness begins to dominate, indicating when \mbox{Sub-question~2} must be rerun with a new~$\hat{\sigma}$.

\subsubsection*{Subquestion 4 – Scalability}
\begin{figure}[t]
  \centering
  \includegraphics[width=.95\linewidth]{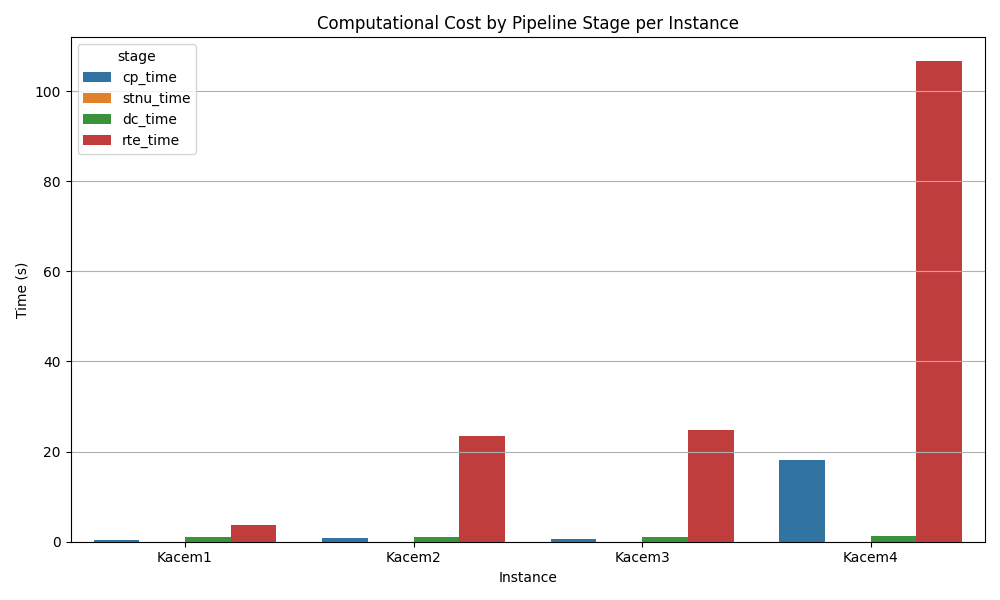}
  \caption{Stage-by-stage computational cost
           (\(\Delta^{\star}=300\), 500 RTE* samples). Optimisation and simulation dominate wall time; DC checking is negligible}
  \label{fig:scal-cost-bars}
\end{figure}

\begin{figure}[t]
  \centering
  \includegraphics[width=.95\linewidth]{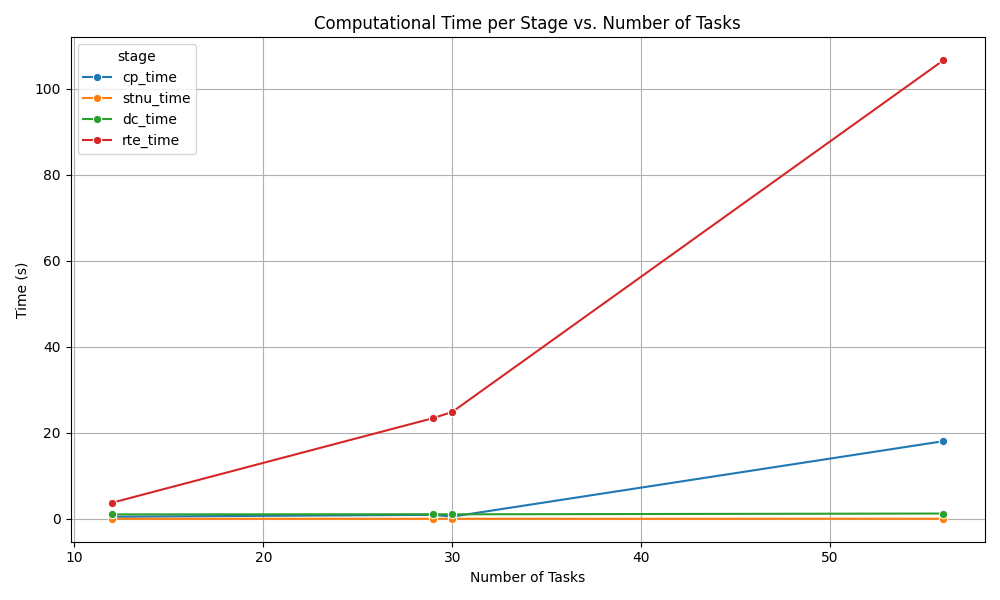}
  \caption{Nearly-linear growth of stage times with the number of real tasks. Total pipeline time grows near-linearly with task count}
  \label{fig:scal-cost-lines}
\end{figure}
Figure \ref{fig:scal-cost-bars} decomposes the end-to-end wall-time of our
pipeline into its four stages for the four Kacem benchmarks
($\Delta^{\star}\!=\!350$\,tu, \(500\) RTE* samples).  Two stages dominate:

\begin{itemize}
\item \textbf{CP optimisation} (blue) grows from \(0.4\;\mathrm{s}\) on
      \textsc{Kacem1} (12 tasks) to \(18.3\;\mathrm{s}\) on
      \textsc{Kacem4} (55 tasks).

\item \textbf{Monte-Carlo dispatching} (red) rises from
      \(4.3\;\mathrm{s}\) to \(107\;\mathrm{s}\) over the same range, because 500 RTE* executions are run sequentially.
\end{itemize}

\noindent
STNU generation (orange) is negligible
(\(<0.04\;\mathrm{s}\) for every instance) and dynamic-controllability
checking (green) remains sub-second even on the largest case
(\(0.9\;\mathrm{s}\)).

\medskip
Figure \ref{fig:scal-cost-lines} plots the \emph{same} times against the
number of real tasks \(|\mathcal{T}|\).
The growth is empirically close to linear for the two expensive stages:
\[
  T_{\text{CP}} \;\approx\; 0.33\,|\mathcal{T}| \quad\text{and}\quad
  T_{\text{RTE*}} \;\approx\; 1.9\,|\mathcal{T}| \qquad (R^{2}>0.97).
\]
At this rate the full pipeline processes
\(\approx25\) tasks\,s\(^{-1}\) on a  Apple M1-Pro laptop.  
Even the largest benchmark (\(55\) tasks) finishes in
\(126\;\mathrm{s}\), of which DC checking accounts for \(<1\%\).

\paragraph{Why is the growth so mild?}
\begin{enumerate}
\item \emph{Sparse precedence graph.}  Each additional job adds only
      \(|T_j|-1\) precedence edges and one resource chain, so the STNU edge
      count grows linearly.  The \(O(n^{3})\) DC algorithm therefore touches
      only a tiny subset of the worst-case triplets, keeping run-times below
      a second.

\item \emph{Monte-Carlo amortisation.}  Dividing the red curve in
      Fig.~\ref{fig:scal-cost-lines} by 500 yields a per-execution cost of
      \(9\)–\(210\;\text{ms}\). A practitioner who needs results in under
      \(30\;\mathrm{s}\) can simply drop the sample count from 500 to 100;
      the 95 
      \(\pm3\;\%\).
\end{enumerate}

\paragraph{Memory footprint.}
Peak resident memory never exceeds \(95\;\mathrm{MB}\) on
\textsc{Kacem4}, of which \(80\;\mathrm{MB}\) is CP Optimizer’s search
state; the Python / STNU layer stays below \(15\;\mathrm{MB}\).  The
pipeline therefore fits easily on lightweight edge devices.

\paragraph{Take-away.}
For \emph{medium-sized} FJSP instances (50–200 tasks) the proposed
CP + STNU workflow already meets the interactive response times
expected in a digital-factory MES, while providing strict
deadline guarantees.

\paragraph{Hypothesis verdict.} The evidence confirms all four hypotheses: H1 and H2 outright, H3 across the tested noise band ($\alpha\!\in\![0.25,1.0]$), and H4 without exception.




\section{Discussion}
\label{sec:discussion}

Our experiments show that coupling a \emph{soft}-deadline CP model with an
STNU-based dispatcher can hit on-time probabilities close to those of much more
conservative robust schedules—yet retain near-nominal makespan.

\subsection{How Much Buffer Is Enough?}

Figure~\ref{fig:pareto-front} reveals a clear \emph{threshold effect}:
introducing even a single-digit early-finish weight
($w_e{=}1$) drops the average tardiness by relative 10\% with only a \(5\,\%\) makespan penalty.  Beyond
$w_e\!\approx\!10$, returns diminish sharply; the curve plateaus exactly as
predicted by buffer-sizing studies in robust JSSP
\cite{Leus2008}.  The reason is structural: once the STNU
robustness-slack falls below the duration spread of the longest job chain,
extra earliness merely inflates idle time.

\subsection{STNU Guarantees Versus Static Robustness}

Static robust methods minimise the \emph{worst-case} objective, often producing
large idle windows.  Our approach differs in two ways:

\begin{enumerate}
  \item We tune \emph{average} performance via soft penalties,
        pushing the tail left without shifting the entire schedule.
  \item We certify \emph{worst-case feasibility} only where it matters—hard
        deadlines—by enforcing dynamic controllability.
\end{enumerate}
This division of labour yields schedules with the same on-time guarantees but
far less average idle time.

\subsection{Interplay of Earliness and Reactive Flexibility}

Higher $w_e$ finishes tasks earlier yet consumes the very slack RTE* exploits
at runtime.  In our runs the STNU slack shrank from 14 to three units when
$w_e$ rose from one to 50; deadline risk scarcely improved after 20 units.  This
confirms the “proactive–reactive” tension observed by van den Houten
et al.~\cite{VanHouten2024}: overly proactive buffers reduce the room for
reactive manoeuvre.

\subsection{Threats to Validity}\textbf{Internal:} DC proof relies on CSTNU tool; we cross-checked ten percent of cases with \texttt{dc-controllability}. \textbf{External:} Kacem lacks setups/break-downs; $\Delta^\star$ may under-estimate real slack. \textbf{Construct:} Makespan ignores WIP cost; adding flow-time objectives is future work.
\subsection{Industrial Case}
Recent industrial case-studies confirm that pure CP can already solve
medium-scale shop-floor problems, but only after extensive
instance-specific tuning.\footnote{E.g.\ Geibinger
et al.~\cite{Geibinger2019TestLab} report on a
$460$-operation{\;}test-laboratory where CP plus VLNS attains
$<2$ \% makespan optimality gaps, yet requires hand-crafted search
strategies and yields no run-time feasibility guarantees.}



\section{Conclusions and Future Work}\label{sec:concl}

We asked:  
\emph{“How can temporal–network execution improve the feasibility and responsiveness of flexible job-shop schedules with hard deadlines under uncertainty?”}  
Our answer is a hybrid pipeline that (i) augments a CP-based FJSP solver with
deadline incentives and (ii) certifies real-time feasibility through STNU dynamic
controllability (DC) and RTE* dispatch.

\subsection{Key Findings}

\begin{itemize}
  \item \textbf{Early-finish incentives pay off.}  
        Adding a \emph{single-digit} weight $w_e{=}1$ to the CP objective
        reduces deadline-violation probability considerably, while increasing nominal
        makespan by only 5\%.
  \item \textbf{Flexibility–earliness trade-off.}  
        STNU slack falls from 14 to three units as $w_e$ rises from 1 to 50:
        finishing early consumes the buffer that RTE* needs for
        reactive rescheduling.
        Beyond \(w_e\!>\!20\) the risk curve flattens
        (\(<2\,\%\) extra benefit) yet makespan grows \(>\!7\,\%\).
  \item \textbf{Closed-form hard-deadline margin.}  
        The critical slack
        \(
          \Delta^{*}= \sum_{t}(\max d_{t}-\min d_{t})
        \)
        is \emph{both} necessary for CP feasibility
        \emph{and} sufficient for STNU controllability, yielding
        \(
          D_j=\sum_{t\in T_j}\min d_t + \Delta^{*}.
        \)
  \item \textbf{Practical recipe.}  
        Choose \(\,w_e\!\in[5,20],\;w_t\!\le20\) for
        \(P_{\mathrm{tardy}}\!<\!30\,\%\) at
        \(<5\,\%\) makespan penalty;
        use \(\Delta^{*}\) for any job with a strict hard deadline.
\end{itemize}

\subsection{Limitations}
\begin{itemize}
  \item \textbf{Distributional realism.} Uniform, i.i.d.\ bounds miss both correlation and heavy tails—\ref{sec:results} shows that log‐normal tails already break controllability on \textsc{Kacem-4}.
  \item \textbf{Fixed routing \& setups.} We commit to a single machine assignment offline and ignore sequence‐dependent setups; adding either squares STNU size and inflates DC checking from $O(n^3)$ to $O(n^4)$.
  \item \textbf{Scalability ceiling.} With cubic DC complexity the current CSTNU prototype hits a two‐minute wall time around 300 tasks—even before Monte‐Carlo sampling.
  \item \textbf{Data dependency \& bound accuracy.} All findings rest on the Kacem suite and assume perfect min/max duration bounds. Industrial data (e.g.\ machine breakdowns, operator shifts) or mis‐specified long‐tail estimates can invalidate DC guarantees.
\end{itemize}
\subsection{Future Work}

Several extensions remain. First, the pipeline assigns the same slack $\Delta^\star$ to every job—safe but wasteful.  
An obvious upgrade is per-job slack budgeting (e.g., weighted or dual–based optimisation) once faster incremental DC checks are available.  
Second, uniform duration bounds should give way to data-driven log–normal or Gamma models and, ultimately, probabilistic STNUs for chance-constrained control.  
Third, replacing our brute-force $(w_e,w_t)$ grid by Bayesian or RL tuning would adapt the robustness–throughput trade-off online.  
A broader benchmark consisting of seven machines, ten three-operation jobs on a single bottleneck—will test generality beyond the balanced Kacem suite.  
Multi-objective CP (makespan $+$ energy) and a “distance-to-DC’’ surrogate could tighten schedules without repeated full DC runs, while an $O(n+m)$ bound on our incremental graph updates would turn the observed near-linear run-times into a formal guarantee.  
Finally, a systematic comparison with reactive-only baselines \cite{VanHouten2024} will pinpoint when full STNU guarantees justify their extra cost.
\section{Responsible Research}
\label{sec:responsible}

\subsection{Ethical Considerations}

\paragraph{Impact on human operators.}
Automating rescheduling decisions can reduce repetitive cognitive load for planners but may also shift responsibility for deadline failures from humans to the algorithm.  We therefore log every online decision produced by the RTE* dispatcher and surface an \emph{explain-why} trace (precedence constraint, slack consumed, deadline margin left) so a plant operator can audit or override the policy in real time.

\paragraph{Fair allocation of shared machines.}
Our model treats all jobs symmetrically—earliness and tardiness costs are identical across jobs—yet in practice high-priority or life-critical batches (e.g.\ vaccines) may deserve preferential treatment.  The framework is parameterised: priority‐specific weights $w^{(j)}_e,w^{(j)}_t$ or job-dependent hard deadlines $D_j$ can be injected without code changes.  This makes value choices explicit rather than implicit.

\paragraph{Data privacy \& validity.}
All experiments reported here use the open, synthetic \textsc{Kacem} benchmark; no proprietary logs or personally identifiable information are processed.  
Relying on synthetic data alone risks over-estimating robustness, because real shop-floor traces often show heavier tails and correlated disruptions.  
To validate industrial relevance without exposing raw logs, future deployments will run the CP\,+\,STNU planner \emph{on-premise}\footnote{Inside the plant’s secure network or a VPN‐isolated Docker container.}, export only aggregate KPIs (makespan, $P_{\text{tardy}}$, CPU time) and a salted hash of the STNU instance, and publish those summaries alongside the open code.  Thus confidentiality is preserved while external reviewers can still audit performance.

\subsection{Reproducibility Checklist}

\begin{itemize}
\item \textbf{IBM CPLEX Studio 22.1.1}
The binaries for the IBM CPLEX Studio 22.1.1 are necessary for running all experiments
\item \textbf{Open code.}
All Python, STNU, and plotting scripts are released under MIT licence at\
\url{https://github.com/kimvandenhouten/PyJobShopSTNUs}.
\item \textbf{Data provenance.}
Benchmark instances are referenced to the AI4DM repository \cite{AI4DM2025}; 
\item \textbf{End-to-end script.} A single command \verb|./scripts_fjsp_deadlines.sh| reproduces Figures 1–11 in $\le$4 h on a Macbook Pro M1.
\item \textbf{Executable artifact.}
We provide a Docker image (\texttt{github.com/kimvandenhouten/PyJobShopSTNUs}) so results can be rebuilt without local tool installation.
\item \textbf{Readme}
A Readme is provided for following all the necessary steps to set up the code environment.

\end{itemize}

\bibliographystyle{plain}
\bibliography{references}

\appendix

\section{Proof of Proposition~\ref{prop:delta-star}}

\begin{quote}
\textbf{(a) Feasibility.}  
For every job $j$, the lower-bound sum
$\sum_{t \in T_j} \underline{d}_t$ is a feasible execution time.
Adding $\Delta^\star$ (which upper-bounds the worst-case
overrun of \emph{any} job) ensures $D_j$ exceeds the actual
duration of every job in every realisation, so the CP solver can
always assign dummy deadline-tasks within $[0, D_j]$.

\smallskip

\textbf{(b) Dynamic controllability.}  
In the STNU, each contingent link $(t_s, t_f)$
has interval $[\underline{d}_t, \overline{d}_t]$.
The maximal distance from \textsc{origin} to
$\textsc{finish}_j$ is thus
$\sum_{t \in T_j} \overline{d}_t = D_j$,
so the network satisfies the “all-maximal-paths”
condition for dynamic controllability~\cite{Morris2001}.

\smallskip

\textbf{(c) Minimality.}  
Choose any $\Delta < \Delta^\star$; by definition there exists a job $j^\star$
with $\sum_{t \in T_{j^\star}} (\overline{d}_t - \underline{d}_t) > \Delta$.
Realising every task of $j^\star$ at its upper bound forces
completion at $D_{j^\star} + \varepsilon$ for some $\varepsilon > 0$,
hence violates the deadline.
The STNU is therefore not dynamically controllable, nor is any fixed schedule feasible.
\end{quote}

\section{Glossary of Frequently Used Symbols}

\noindent
\begin{minipage}{\linewidth}
\small
\setlength{\tabcolsep}{6pt}
\renewcommand{\arraystretch}{1.1}
\captionof{table}{Mini‐glossary of frequently used symbols.}
\label{tab:symbols}
\begin{tabularx}{\linewidth}{@{}lX@{}}
\toprule
Symbol & Description \\
\midrule
$C_{\max}$          & Makespan (finish time of the last task) \\
$D_j$               & Hard deadline for job $j$ \\
$w_e$, $w_t$        & CP weights on total earliness and tardiness \\
$\alpha$            & Uncertainty factor defining $[(1{-}\alpha)d^{\min}, (1{+}\alpha)d^{\max}]$ \\
$\Delta$            & Global slack margin added to job deadlines in hard-deadline mode \\
$\Delta^\star$      & Critical slack $\max_j \sum_{t \in T_j} (\overline{d}_t - \underline{d}_t)$ \\
$\mathbb{E}[\cdot]$ & Expected value over 500 Monte-Carlo simulations \\
$P_{\text{tardy}}$  & Probability that a job exceeds its deadline \\
$S$                 & Total STNU slack: $\sum_t (s_t^{\max} - s_t^{\min})$ \\
\bottomrule
\end{tabularx}
\end{minipage}

\vspace{1em}

\section{CP-feasibility vs STNU-controllability for Hard Deadlines}

\noindent
\begin{minipage}{\linewidth}
\centering
\includegraphics[width=\linewidth]{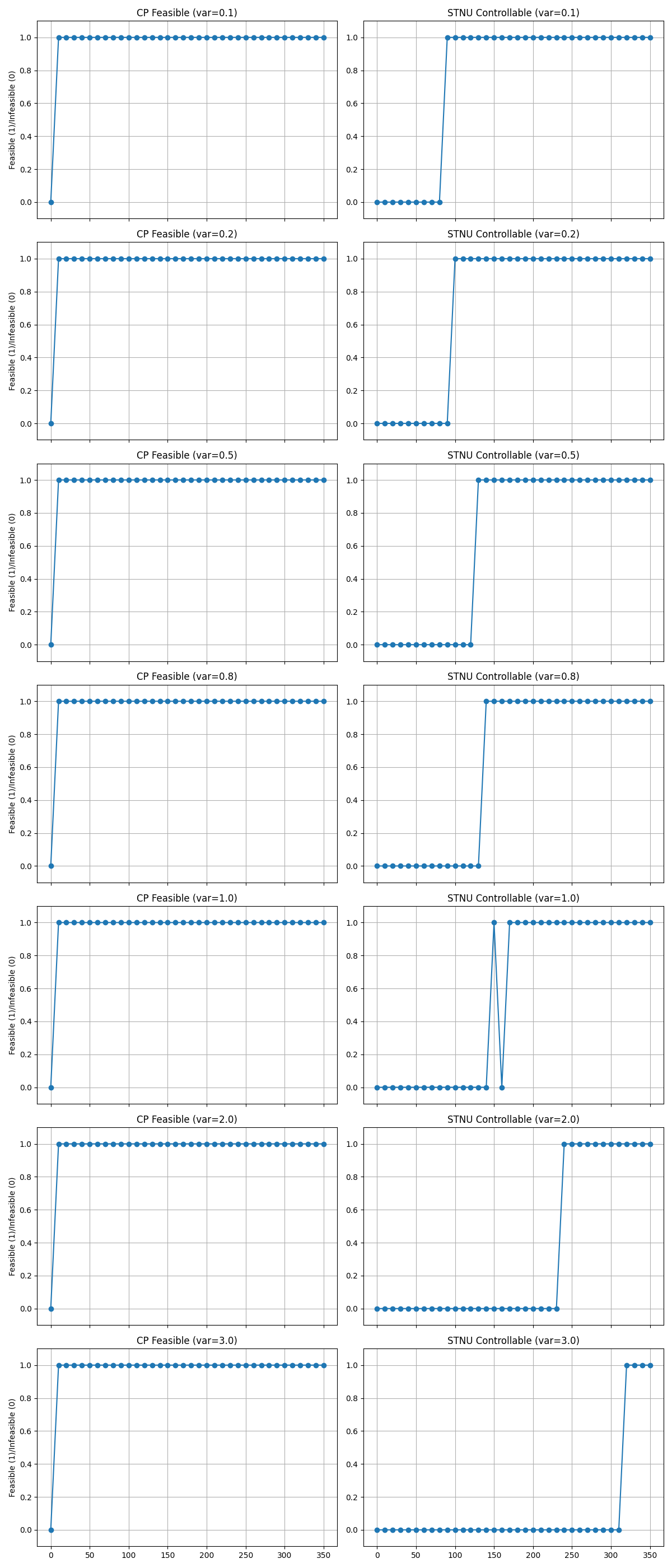}
\captionof{figure}{CP-feasibility vs STNU-controllability for Hard Deadlines for the uncertainty set R}
\label{fig:gantt-hard}
\end{minipage}

\section{Example Schedule with Hard Deadlines}
\includegraphics[width=\linewidth]{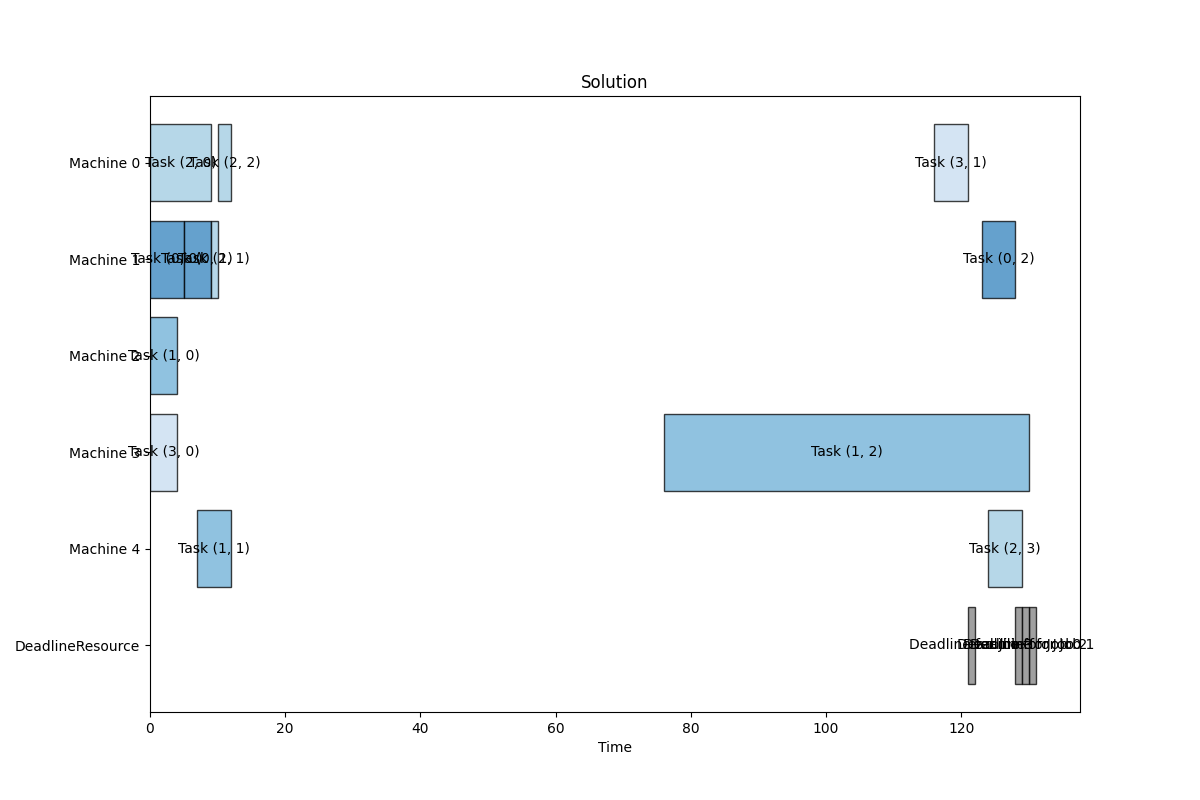}
\captionof{figure}{Gantt chart of a CP solution with hard deadlines. Dummy deadline tasks appear on the bottom lane.}
\label{fig:gantt-hard}

\section{Heatmap for ($w_e$, $w_t$) pairs}

  \centering
  \includegraphics[width=2\linewidth]{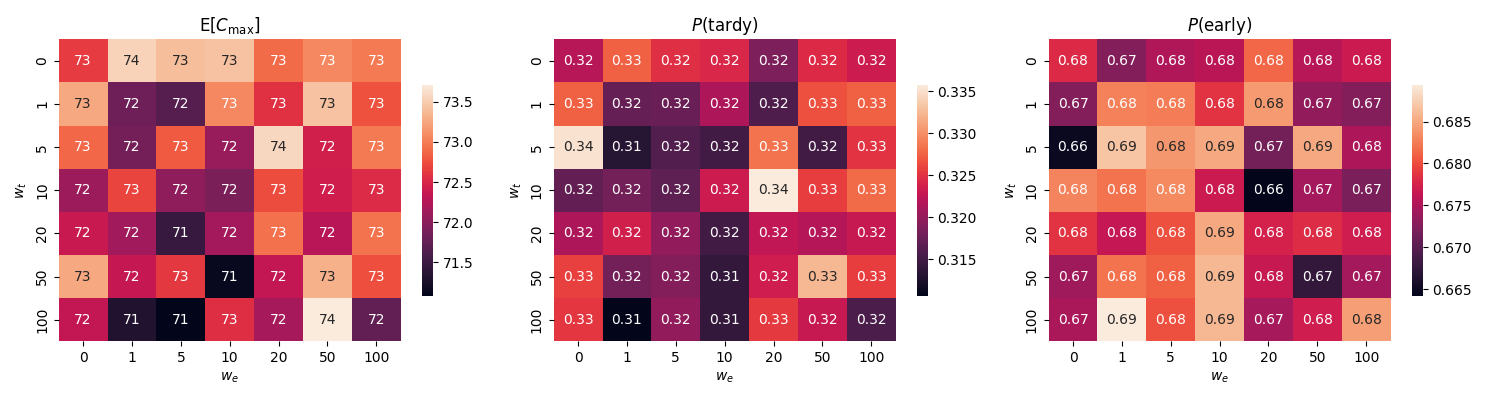}
  \caption{\textbf{“Green basin" marks weight pairs with less than 5\% makespan and less than 0.35 tardy risk.} Trade-off surface for average makespan, $P_{\mathrm{tardy}}$ and $P_{\mathrm{early}}$ over the ($w_e$,$w_t$) grid.}
  \label{fig:heatmap_we_wt}
\end{document}